\documentclass[letterpaper]{article} 
\usepackage{aaai23}  
\usepackage{times}  
\usepackage{helvet}  
\usepackage{courier}  
\usepackage[hyphens]{url}  
\usepackage{graphicx} 
\usepackage{natbib}  
\usepackage{caption} 
\usepackage{booktabs}
\usepackage{algorithm}
\usepackage{algorithmic}
\usepackage{amsmath}
\usepackage{bm}
\usepackage{multirow}
\usepackage{float}
\usepackage{pifont}
\usepackage{url}
\usepackage{caption}
\usepackage{stfloats}

\usepackage{newfloat}
\usepackage{listings}
\DeclareCaptionStyle{ruled}{labelfont=normalfont,labelsep=colon,strut=off} 
\floatstyle{ruled}
\newfloat{listing}{tb}{lst}{}
\floatname{listing}{Listing}
\title{Tagging before Alignment: Integrating Multi-Modal Tags for Video-Text Retrieval}
\author{
    Yizhen Chen,\textsuperscript{\rm 1}
    Jie Wang,\textsuperscript{\rm 1}
    Lijian Lin,\textsuperscript{\rm 2}
    Zhongang Qi,\textsuperscript{\rm 2}\thanks{Corresponding author.}
    Jin Ma,\textsuperscript{\rm 1}
    Ying Shan\textsuperscript{\rm 2}
}
\affiliations{
    \textsuperscript{\rm 1}IPS Search, Tencent PCG\\
    \textsuperscript{\rm 2}ARC Lab, Tencent PCG\\
    \{grayyzchen, zhongangqi, yingsshan\}@tencent.com, photonicsjay@163.com, ljlin@stu.xmu.edu.cn, majin01@mail.ustc.edu.cn

}

\usepackage{bibentry}

\begin{document}

\maketitle

\begin{abstract}
Vision-language alignment learning for video-text retrieval arouses a lot of attention in recent years. Most of the existing methods either transfer the knowledge of image-text pretraining model to video-text retrieval task without fully exploring the multi-modal information of videos, or simply fuse multi-modal features in a brute force manner without explicit guidance. 
In this paper, we integrate multi-modal information in an explicit manner by tagging, and use the tags as the anchors for better video-text alignment. Various pretrained experts are utilized for extracting the information of multiple modalities, including object, person, motion, audio, etc. 
To take full advantage of these information, we propose the TABLE (TAgging Before aLignmEnt) network, which consists of a visual encoder, a tag encoder, a text encoder, and a tag-guiding cross-modal encoder for jointly encoding multi-frame visual features and multi-modal tags information. 
Furthermore, to strengthen the interaction between video and text, we build a joint cross-modal encoder with the triplet input of [vision, tag, text] and perform two additional supervised tasks, Video Text Matching (VTM) and Masked Language Modeling (MLM). 
Extensive experimental results demonstrate that the TABLE model is capable of achieving State-Of-The-Art (SOTA) performance on various video-text retrieval benchmarks, including MSR-VTT, MSVD, LSMDC and DiDeMo. 
\end{abstract}

\section{Introduction}

Vision-language alignment learning for video-text retrieval becomes an emerging requirement with the increasing of videos and short videos uploaded online, and has attracted great attention in recent years. In the research field, it returns the most relevant videos for a given text query to facilitate large-scale videos searching and management. In the recommendation field, it recommends the most relevant text queries for the video that the user is watching to prompt more searching and browsing.
Although a lot of recent works \cite{zhu2020actbert, gabeur2020multi, dzabraev2021mdmmt, lei2021less, liu2021hit, luo2021clip4clip, fang2021clip2video, cheng2021improving, chen2021mind, cao2022visual} have made remarkable progress, the cross-modal alignment between video and text remains a challenging task.

\begin{figure}[t]
	\centering
	\includegraphics[width=\linewidth]{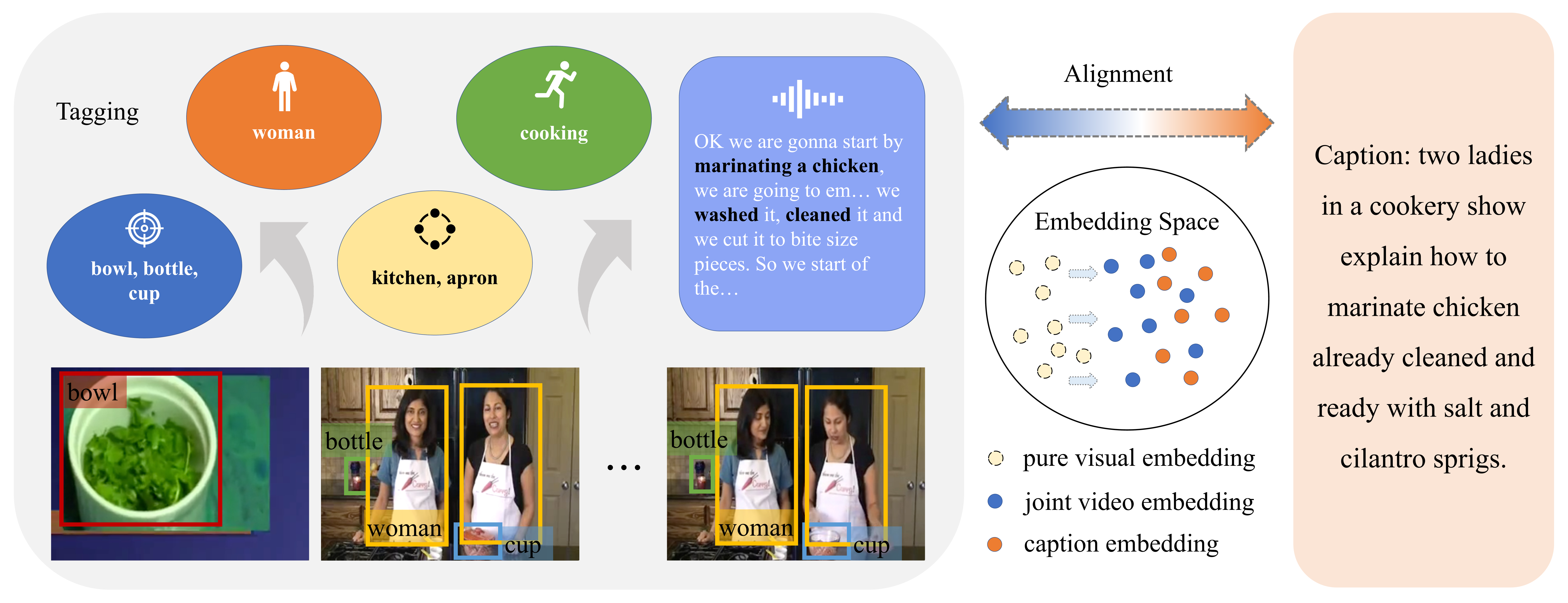}
	\captionsetup{font={small}}
	\caption{The multi-modal information of a video can be extracted and converted to explicit tags to motivate the video-text alignment. As shown, with multi-modal tags, the joint video embedding is closer to caption embedding than pure visual embedding.}
	\label{fig1}
\end{figure}

Most of the existing video-text retrieval models use multi-layer transformers to learn generic representations from massive video-text pairs, which can be roughly divided into two categories. 
The first category uses only frame features to transfer the knowledge of image-text pretrained model to video-text retrieval task without fully exploring the multi-modal information of videos \cite{lei2021less, luo2021clip4clip, fang2021clip2video, cheng2021improving, cao2022visual}.  
A representative method is CLIP4Clip \cite{luo2021clip4clip}, which utilizes the knowledge of the CLIP (Contrastive Language-Image Pretraining) \cite{radford2021learning} model to visually encode multi-frame information as an overall representation of the video. The model achieves good performance on multiple benchmarks, demonstrating the effectiveness of the knowledge transferring.
The second category is to jointly encode the information of various modalities in the video, such as objects, actions, scenes, audio, etc. \cite{gabeur2020multi, dzabraev2021mdmmt, wang2021t2vlad, hao2021multi}.
A typical representative method is MMT (Multi-Modal Transformer) \cite{gabeur2020multi}, which uses multiple pretrained experts to extract the embeddings of different modalities of the video, and builds a multi-modal transformer for feature fusion. 
The performance of this model demonstrates the advantage of using multi-modal information over using only single visual information of the video.
However, most of these methods simply fuse multi-modal features in a brute force manner without explicit guidance, which increases the difficulty of the learning process. 

In this paper, we introduce a novel approach which not only utilizes image-text pretrained model for knowledge transfer, but also fully exploits the multi-modal information of video in an explicit manner to guide the visual-language alignment.
A set of pretrained experts are utilized to extract the features from a diversity of modalities of videos, including object, person, scene, motion, and audio. Object and person focus on instance features or local features of the video; scene focuses on background or global features of the video. The motion features reveal the behavior or action happening in the video, which extract temporal information from consecutive frames. The audio provides supplementary descriptions of the video, which may not appear in visual features. The most challenging part of the proposed approach is how to fuse the features of different modalities effectively and efficiently. In this work, we utilize tagging as the bridge, whose advantages lie in the following two aspects. First, embeddings from various modalities generated by different experts are not compatible from each other; while tagging generates unified representations for different modalities. Second, multi-modal tags can be treated as anchor points to make the visual-language alignment learn the local feature, the global feature, the temporal feature, and other supplementary features comprehensively, which eases the learning process significantly.
For example, in Fig. \ref{fig1}, through object detector, we can identify objects appearing in the video, and label them as {\it bowl}, {\it bottle}, and {\it cup}; the people in the video can also be precisely identified as {\it woman}. Through image classification on multiple frames, we can obtain the scene tag, i.e., {\it kitchen}, and some supplementary tags, like {\it apron}.
Through action recognition, we can recognize the ongoing behavior in the video and get the motion tag, i.e., {\it cooking}.
Further, we can convert the audio of the video to text by Automatic Speech Recognition (ASR), and extract the keywords from it as audio tags, i.e., {\it marinating a chicken}, {\it washed}, and {\it cleaned}. 
The above tags provide abundant information from multi-modality of the video, which bridge the gap between vision and language, making the alignment learning more efficiently and precisely.

Specifically, we construct a TABLE (TAgging Before aLignmEnt) network, which utilizes a visual encoder to transfer the knowledge of image-text pretrained model to extract the frame features, a tag encoder to extract information of multi-modal tags, and a cross-modal encoder to jointly learn multi-frame visual features and multi-modal tags features of the video.
The multi-modal tags are used as anchors to guide the visual-text alignment. The temporal information between frames is also learned by the cross encoder.
Further, we form a triplet input consisting of [vision, tag, text] to perform two auxiliary tasks, namely Video Text Matching (VTM) and Masked Language Modeling (MLM). 
The VTM task treats the vision and tag information as a whole, and trains the model to identify whether it matches the text query. 
We perform in-batch hard negative mining by the contrastive similarity distribution. 
The objective of the MLM task is to recover the randomly masked words according to the triplet of [vision, tag, unmasked text]. 

To summarize, the contributions of this work lie in four-fold:  
\begin{itemize}
	
	\item We propose a novel method for video-text retrieval, which not only transfers image-text knowledge, but also fully exploits the multi-modal information of video, including object, person, scene, motion and audio.
	
	\item We integrate multi-modal information by an explicit and unified method, tagging, and use them as anchors for guiding visual-text alignment.
	
	\item We build a TABLE network to jointly encode multi-frame visual features and multi-modal tag information. And we introduce VTM and MLM as auxiliary supervisions to strengthen video-text interaction.
	
	\item The proposed TABLE model achieves SOTA performance on several video-text retrieval benchmarks, including MSR-VTT, MSVD, LSMDC and DiDeMo. 
	
\end{itemize}

\begin{figure*}[h]
	\centering
	\includegraphics[width=0.7\linewidth]{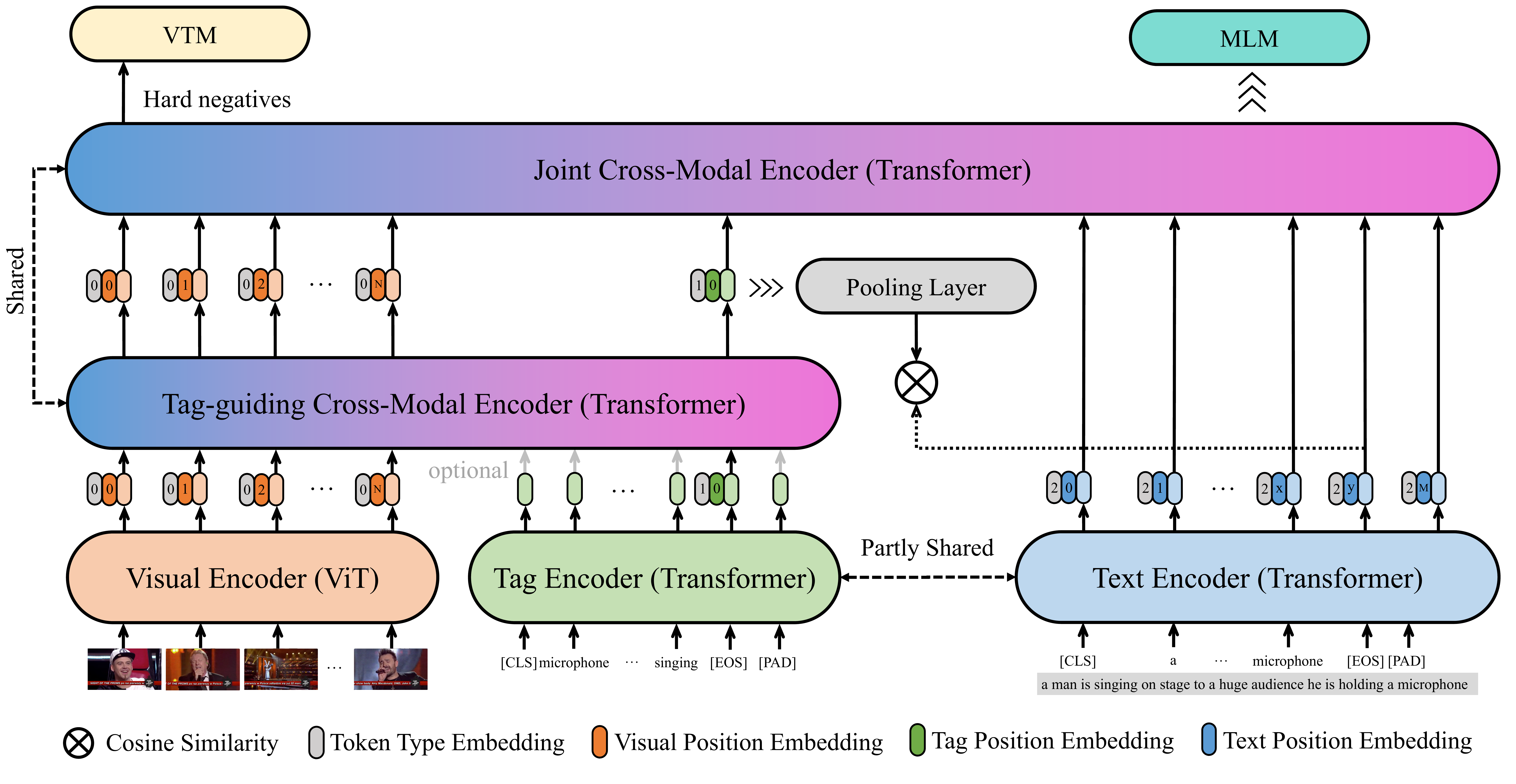}
	\captionsetup{font={small}}
	\caption{Overall framework of our proposed TABLE (TAgging Before aLignmEnt) model.}
	\label{fig2}
\end{figure*}

\section{Related Work}

\subsection{Video representation learning} 
A lot of approaches have been proposed for video representation learning, which can be classified into convolution-based methods and transformer-based methods.
Many previous works \cite{tran2015learning, xie2018rethinking, feichtenhofer2019slowfast} employed 2D or 3D convolutional network to encode spatial and temporal information of the videos. 
Recently, the Vision Transformer (ViT) was proposed and attained excellent results on image classification compared to convolutional methods. 
Since then, many recent works applied ViT to encode the visual features of videos, such as ViViT \cite{arnab2021vivit} and TimeSformer \cite{bertasius2021space}, which generally decoupled the spatial and temporal information by designing two-stage transformer structures.
Most of the above approaches exploited temporal information by improving image representation, however, the abundant multi-modal information of videos were not fully utilized.  
Our work performs temporal encoding with cross-modal interaction and  exploits comprehensive information from multiple modalities,  which are extracted by various pretrained experts, including object detection \cite{ultralytics2020yolov5}, action recognition \cite{xie2018rethinking}, speech recognition \cite{stewart2013robust}, etc. 

\subsection{Video-Text Retrieval} 
Recent video-text retrieval approaches can be divided into two types according to whether the multi-modal information of videos is utilized or not. The first type simply adopted the visual features for video-text alignment,
where CLIP-based methods \cite{luo2021clip4clip, fang2021clip2video, cheng2021improving, cao2022visual} show obvious advantages in recent years. 
CLIP4Clip \cite{luo2021clip4clip} firstly transferred the knowledge of large-scale image-text pretraining to the task of video-text retrieval with fine-tuning. 
Based on it, CLIP2VIDEO \cite{fang2021clip2video} proposed a temporal different block to capture video’s motion feature, and a temporal alignment block to re-align the tokens of video clips and phrases.
The second type used multi-modal information contained in videos for enhancing video-text alignment \cite{gabeur2020multi, dzabraev2021mdmmt, wang2021t2vlad, hao2021multi}. 
MMT \cite{gabeur2020multi} exploited multi-modal information extracted by seven pretrained experts but only fused them in a brute force manner without explicit guidance. MDMMT \cite{dzabraev2021mdmmt} emphasized the importance of the motion information on the basis of MMT. 
Wang et al. \cite{wang2021t2vlad} and Hao et al. \cite{hao2021multi} performed local alignment between multi-modal features and text features, but the gap between the embeddings of different modalities is too huge to bridge.
To solve this problem, we propose to integrate multi-modal information by an explicit method, i.e., tagging.
Tagging generates unified representations to erase the gap between different modalities, and can be treated as anchors to guide the visual-language alignment more explicitly. 
Some image-text retrieval methods \cite{zhen2019deep, qian2021dual} also used tag to boost performance, but our work is different in many ways. Firstly, these work only used the unimodal tag of image, while our work fully extracts the multi-modal tags of video, which is more comprehensive. Secondly, these work used classification loss as extra supervision, while our method does not directly use tag as training target, but regards it as anchor for video-text alignment.

\section{Methodology}
Given a set of videos and texts, our goal is to learn accurate representations and calculate their similarities. The video (or text) candidates are then ranked by their similarities with the query text (or video) in the field of text-to-video (or video-to-text) retrieval. 
To achieve this goal, video and text need to be aligned in a joint embedding space, which is difficult due to the independence of the two feature extraction processes.  
Therefore, we propose to fully employ the multi-modal tags of the video and utilizes them as anchors to explicitly motivate the semantic alignment between visual and text. Specifically, we construct a TABLE (TAgging Before aLignmEnt) network to jointly encode multi-frame visual information and multi-modal tag information, meanwhile the temporal information is also modeled. 
Moreover, we incorporate additional Video Text Matching (VTM) and Masked Language Modeling (MLM) losses. The training target of VTM is to judge whether the text matches the fusion of visual and tag features. The objective of MLM is recovering the masked text word according to video, tag and the context information. 
As shown in Fig. \ref{fig2}, our model consists of a visual encoder, a tag encoder, a text encoder, a tag-guiding (\textbf{TG}) cross-modal encoder and a joint cross-modal encoder. The two  cross-modal encoders share paramteters with each other and performs multi-modal information integration.

\subsection{Multi-modal Tag Mining}
In this paper, multiple pretrained experts are used to extract related tag information of individual modality. Firstly, we adopt the yolov5s \cite{ultralytics2020yolov5} model pretrained on the COCO \cite{lin2014microsoft} dataset for video object detection. Main objects appearing in multiple frames with high confidence score are selected with object tags \bm{$t_{obj}$}. 
In particular, we choose the detection model pretrained on Open Image dataset \cite{kuznetsova2020open} to distinguish person into man, woman, boy and girl, obtaining specific person tags \bm{$t_{per}$}. 
Secondly, we employ the COCO pretrained ViT-B-16 \cite{dosovitskiy2020image} model to perform image classification of video frames, where predicted categories with high confidence score are regarded as the scene tags \bm{$t_{sce}$}.
Object and person tags focus on instance or local features, while scene tags put attention to global information of video frames.
Thirdly, we adopt the S3D \cite{xie2018rethinking} network pretrained on Kinetics-400 dataset \cite{carreira2017quo} to obtain the motion tags \bm{$t_{mot}$}, which is important for video understanding.
Furthermore, we apply the Automatic Speech Recognition (ASR) API of iFLYTEK to get video transcripts and then employ the KeyBert \cite{grootendorst2020keybert} to extract transcript keywords as audio tags \bm{$t_{aud}$}. 
The audio modality usually provides complementary information which is always not included in visual features. At last, we concatenate the above tags and obtain the whole multi-modal tag information of the video, \bm{$t_{mul}}=[\bm{t_{obj}}, \bm{t_{per}}, \bm{t_{sce}}, \bm{t_{mot}}, \bm{t_{aud}$}].

\subsection{Visual and Text Encoder}
\textbf{Visual Encoder.} 
We uniformly sample $N$ frames to form a sequence as the video representation,  $f_i=\left\{f_i^1, f_i^2, ..., f_i^N\right\}$.
Inspired by the successful transferring of the image-text pretraining knowledge to video-text learning, we directly adopt the ViT model in CLIP \cite{radford2021learning} to extract visual features, which processes non-overlapping image patches of individual frame and  linearly project them to 1D token sequence. 
The patch tokens are then passed through a 12-layer transformer to realize self-attention procedure. The [CLS] embedding is projected into a normalized lower-dimensional embedding space with a MLP layer to obtain the overall representation of each frame. 
For simple presentation, we omit the process of taking [CLS] token of ViT in Fig. \ref{fig2}, and thus the output of the visual encoder is the sequential representations of multiple frames, \bm{$v_i}=\left\{\bm{v_i^1}, \bm{v_i^2}, ..., \bm{v_i^N}\right\}$.

\textbf{Text Encoder.} We apply the BERT-base encoder in CLIP \cite{radford2021learning} to obtain tag and caption embedding, which is a 12-layer transformer with 512 dimensional width and 8 attention heads. The transformer output of tag and caption can be expressed as \bm{$t_i}=\left\{\bm{t_i^1}, \bm{t_i^2}, ..., \bm{t_i^K}\right\}$ and \bm{$c_i}=\left\{\bm{c_i^1}, \bm{c_i^2}, ..., \bm{c_i^M}\right\}$ respectively, where $K$ and $M$ represents the token length of tag and caption. The [EOS] embedding of the last layer is passed through a linear projection layer and regarded as the overall representation of the text (e.g., tag and caption). Particularly, the parameters of transformer blocks are shared between tag and caption encoders, except the linear projection layer. 

\subsection{Tag-guiding cross-modal encoder}
To bridge the visual-text semantic gap for alignment, we introduce multi-modal tags as anchors. As shown in Fig. \ref{fig2}, we concatenate multi-frame visual features and multi-modal tag embeddings as the input of the cross-modal encoder.
Specifically, the concatenated input can be represented by $\left\{\bm{v_i^1}, \bm{v_i^2}, ..., \bm{v_i^N}; \bm{t_i^e}\right\}$, where $\bm{t_i^e}$ represents the [EOS] embedding of the tag encoder output and the overall information of the multi-modal tags.
The \textbf{TG} cross-modal encoder is composed of 4-layer transformer with 512 dimensional width and 8 cross-attention heads. 
The position embedding and transformer parameters are initialized by the weights of first four layers of the CLIP’s text encoder. The \textbf{TG} cross-modal encoder conducts deep fusion of multi-frame visual features with multi-modal tag embeddings, meanwhile modelling the temporal information of the video. The fused output can be expressed as  $\left\{\bm{g_i^1}, \bm{g_i^2}, ..., \bm{g_i^N}; \bm{g_i^{e}}\right\}$. 
We then apply an average pooling layer to acquire the overall representation, which can be formulated as  $\bm{g_i^o}=\bm{\rho}(\left\{\bm{g_i^1}, \bm{g_i^2}, ..., \bm{g_i^N}; \bm{g_i^{e}}\right\})$.
In addition, to fully exploit the pretrained image-text knowledge, we adopt residual connection between the pooled visual features and the overall cross-modal representation, which can be represented by $\bm{\hat{g}_i^o}=\lambda\bm{g_i^o}+ \bm{\rho}(\left\{\bm{v_i^1}, \bm{v_i^2}, ..., \bm{v_i^N}\right\})$. $\lambda$ is a learnable weight factor. 

The output of text encoder can be denoted by \bm{$c_i}=\left\{\bm{c_i^1}, \bm{c_i^2}, ..., \bm{c_i^N}\right\}$, where the [EOS] embedding $\bm{c_i^e}$ is selected as the overall representation of caption. 
We define a similarity function as $s(V_i, T_i) = \phi(\bm{\hat{g}_i^o})^T \psi(\bm{c_i^e}) $,  where $\phi(\cdot)$ and $\psi(\cdot)$ are the linear projection functions that project visual and caption embedding into the shared semantic space. Then, we construct in-batch similarity between video and text:
\begin{equation}
	\label{tv_loss}
	L_{t2v} = -\frac{1}{B}\sum_{i}^{B}log\frac{exp(s(V_i, T_i)/\tau)}{\sum_{j=1}^{B}exp(s(V_j, T_i)/\tau)},
\end{equation}
\begin{equation}
	\label{vt_loss}
	L_{v2t} = -\frac{1}{B}\sum_{i}^{B}log\frac{exp(s(V_i, T_i)/\tau)}{\sum_{j=1}^{B}exp(s(V_i, T_j)/\tau)},
\end{equation}
\begin{equation}
	\label{total_loss}
	\centering
	L_{con}=\frac{1}{2}(L_{t2v}+L_{v2t}),
\end{equation}
where $\tau$ is a learnable temperature parameter, $B$ is batch size, $L_{t2v}$ and $L_{v2t}$ represents the text-to-video loss and video-to-text loss. $L_{con}$ is the total contrastive loss.

\subsection{Video Text Matching and Masked Language Modeling}
\label{s3.4}
In addition to contrastive loss for visual-text alignment, Video Text Matching (\textbf{VTM}) and Masked Language Modeling (\textbf{MLM}) are incorporated to motivate fine-grained interaction between video and text. 

\textbf{VTM} determines whether the pair of [visual, tag] and [text] is matched, where [visual, tag] is the \textbf{TG} cross-modal fusion output of multi-frame visual features and multi-modal tag information, [text] is the overall representation of text. 
This pair is then fed into a joint cross-modal encoder, which shares parameters with the \textbf{TG} cross-modal encoder. The first output token embedding is regarded as the joint video-text representation $V_i$, and passed through a fully connected layer for binary prediction (match or not match with the text representation $T_i$).  Defining the precition as $p^{vtm}(V_i,T_i)$, the VTM loss is:

\begin{equation}
	\label{vtm_loss}
	L_{vtm} = - \frac{1}{O}\sum_{i}^{O}\sum_{t=0}^{1}y_{it}^{vtm}log_2(p_t^{vtm}(V_i, T_i)),
\end{equation}
where $y_{it}^{vtm}$is a sign function which has the value of 1 if $t$=1 else 0 if $t=0$. $(V_i, T_i)$ is a positive pair when $t$=1, and a negative pair otherwise. $p_t^{vtm}(V_i, T_i)$ denotes the prediction probability of $t$.
$O$ is the total number of video-text pairs for VTM task, which is composed of positive and negative pairs.
We perform hard negative mining strategy when constructing the triplet sample. For each [visual, tag] in mini-batch, we sample a negative [text] according to the in-batch similarity matrix calculated in Equation \ref{tv_loss}.
Likewise, for each [text], we sample a negative [visual, tag] in the mini-batch.

\begin{table*}[ht]
	\centering
	\resizebox{0.8\textwidth}{0.1\textheight}{
		\begin{tabular}{ccc|ccccc|ccccc}
			\toprule
			\ & \  & \ & \multicolumn{5}{c|}{Text-to-Video (T2V)} &  \multicolumn{5}{c}{Video-to-Text (V2T)}\\
			\midrule
			Type & Method & Pretrained Dataset & R@1 & R@5 & R@10 & MdR & MnR & 
			R@1 & R@5 & R@ 10  & MdR & MnR \\
			\midrule
			NO-CLIP & JSFusion \cite{yu2018joint} &- & 10.2 & 31.2 & 43.2 & 13.0 & - & - & - & -  & - & - \\
			NO-CLIP & HT-Pretrained \cite{miech2019howto100m}  &HowTo100M \cite{miech2019howto100m}  & 14.9 & 40.2 & 52.8 & 9.0 & - & - & - & -  & - & - \\
			NO-CLIP & CE \cite{liu2019use}  &-  & 20.9 & 48.8 & 62.4 & 6.0 & 28.2 & 
			20.6 & 50.3 & 64.0  & 5.3 & 25.1 \\
			NO-CLIP & MMT-Pretrained \cite{gabeur2020multi} &HowTo100M  & 26.6 & 57.1 & 69.6 & 4.0 & 24.0 & 
			27.0 & 57.5 & 69.7  & 3.7 & 21.3 \\
			NO-CLIP & TACo \cite{yang2021taco}  &HowTo100M   & 26.7 & 54.5 & 68.2 & 4.0 & - & - & - & -  & - & - \\
			NO-CLIP & SUPPORT-SET \cite{patrick2020support}  &- & 27.4 & 56.3 & 67.7 & 3.0 & - & 
			26.6 & 55.1 & 67.5 & 3.0 & - \\
			NO-CLIP & FROZEN \cite{bain2021frozen}  &WebVid-2M\cite{bain2021frozen}  & 31.0 & 59.5 & 70.5 & 3.0 & - & - & - & -  & - & - \\
			NO-CLIP & HIT-Pretrained \cite{liu2021hit}  &HowTo100M & 30.7 & 60.9 & 73.2 & 2.6 & - & 
			32.1 & 62.7 & 74.1  & 3.0 & - \\
			\hline\hline
			CLIP-based & CLIP \cite{portillo2021straightforward}  &WIT \cite{radford2021learning}  & 31.2 & 53.7 & 64.2 & 4.0 & - & 
			27.2 & 51.7 & 62.6  & 5.0 & - \\
			CLIP-based & MDMMT \cite{dzabraev2021mdmmt}  &WIT+HowTo100M  & 38.9 & 69.0 & 79.7 & 2.0 & 16.5 & 
			- & - & -  & - & - \\
			CLIP-based & CLIP4Clip-meanP \cite{luo2021clip4clip} &WIT  & 43.1 & 70.4 & 80.8 & 2.0 & 16.2 &  43.1 & 70.5 & 81.2  & 2.0 & 12.4 \\
			CLIP-based & CLIP4Clip-seqTransf \cite{luo2021clip4clip}  &WIT  & 44.5 & 71.4 & 81.6 & 2.0 & 15.3 & 42.7 & 70.9 & 80.6  & 2.0 & 11.6 \\
			CLIP-based & VCM \cite{cao2022visual}  &WIT  & 43.8 & 71.0 & 80.9 & 2.0 & 14.3 & 
			45.1 & 72.3 & 82.3  & 2.0 & 10.7 \\
			CLIP-based & CLIP2VIDEO \cite{fang2021clip2video}  &WIT   & 45.6 & 72.6 & 81.7 & 2.0 & 14.6 & 43.5 & 72.3 & 82.1  & 2.0 & 10.2 \\
			CLIP-based & CAMoE \cite{cheng2021improving}  &WIT  & 44.6 & 72.6 & 81.8 & 2.0 & 13.3 & 
			45.1 & 72.4 & 83.1  & 2.0 & 10.0 \\
			CLIP-based & \textbf{TABLE (ours)}  &WIT  & 47.1 & 74.3 & 82.9 & 2.0 & 13.4 & 47.2 & 74.2 & 84.2 & 2.0 & 11.0 \\
			\hline\hline
			CLIP-based & CAMoE$\ast$ \cite{cheng2021improving}  &WIT & 47.3 & 74.2 & 84.5 & 2.0 & 11.9 & 49.1 & 74.3 & 84.3  & 2.0 & 9.9 \\
			CLIP-based & CAMoE-online$\ast$ \cite{cheng2021improving}  &WIT   & 48.8 & 75.6 & \bf{85.3} & 2.0 & 12.4 & 
			50.3 & 74.6 & 83.8  & 2.0 & \bf{9.9} \\
			CLIP-based & \textbf{TABLE$\ast$ (ours)}  &WIT   & \bf{52.3} & \bf{78.4} & 85.2 & \bf{1.0} & \bf{11.4} & \bf{51.8} & \bf{77.5} & \bf{85.1}  & \bf{1.0} & 10.0 \\
			\bottomrule
	\end{tabular}}
	\captionsetup{font={small}}
	\caption{Retrieval results on the MSR-VTT dataset. $\ast$ indicates methods with inference strategy (as in subsequent tables).}
	\label{tab2}
\end{table*}

\textbf{MLM} predicts the masked word based on the [visual, tag] representations and the unmasked context, which is also a classification task. To be specific, we randomly mask the text tokens with a probability of 15\%, and the replacement is the [MASK] token with an 80\% probability, a random token with a 10\% probability, original token with a 10\% probability. 
Supposing $\hat{T}_i$ denotes the masked text and $p^{mlm}(V_i,\hat{T}_i)$ denotes the prediction result of the masked word, the loss function of MLM can be expressed as: 

\begin{equation}
	\label{mlm_loss}
	L_{mlm} = -\frac{1}{Q} \sum_{i}^{Q}\sum_{v=1}^{V}y_{iv}^{mlm}log_2(p_v^{mlm}(V_i,\hat{T}_i)),
\end{equation}
where $y_{iv}^{mlm}$ is a sign function which has the value of 1 if the masked word of sample $i$ is $v$. $V$ is the vocabulary size and $Q$ is the total number of video-text pairs for MTM task.
Finally, the overall loss function of the TABLE model is:
\begin{equation}
	\label{model_loss}
	\centering
	L=L_{con} + L_{vtm} + L_{mlm}
\end{equation}

\subsection{Inference Strategy}
\label{s3.5}
During the inference stage, the joint cross-modal encoder with VTM and MLM loss is discarded, and the similarity confidences between video and text are only calculated by the \textbf{TG} encoder and the text encoder. To further improve the performance, we also adopt the inference strategy proposed in CAMoE \cite{cheng2021improving}, which conduct SoftMax operation to revise the similarity matrix.

\section{Experiments}

\subsection{Datasets}

\subsubsection{MSR-VTT} \cite{xu2016msr} dataset is widely studied in video-text retrieval task, which contains 10,000 videos of 10-32 seconds, each corresponding to 20 captions. We follow the data split in previous method \cite{gabeur2020multi}, i.e., 9000 videos for training and 1,000 videos for testing.

\subsubsection{MSVD} \cite{chen2011collecting} dataset contains 1,970 videos, each video has nearly 40 titles. Train, validation and test sets have 1,200, 100 and 670 videos respectively. Following previous works \cite{luo2021clip4clip}, we report the result on testing set with multiple captions per video. The limited training data makes the learning on this dataset challenging.

\subsubsection{LSMDC} \cite{rohrbach2015long} dataset consists of 118,081 short videos clips extracted from 202 movies. Similar to previous work \cite{luo2021clip4clip}, we validate the performance on the test set containing 1,000 videos. 

\subsubsection{DiDeMo} \cite{anne2017localizing} dataset involves about 10,000 videos and each video has about 4 annotated sentences. 
Following previous works \cite{luo2021clip4clip}, we conduct video-paragraph retrieval task on the test split (with 1,004 videos) by concatenating all sentences of each video into a single query. The challenge of this dataset lies in the alignment of long videos and long texts.

\subsection{Metrics}
We evaluate the model performance with standard retrieval metrics, i.e., Recall at rank K (R@K, K=1,5,10), Median Rank (MdR) and Mean Rank (MnR). R@K represents the proportion of the ground-truth result included in the top-K recalled results. MdR and MnR represent the median and mean rank of correct results respectively. Therefore, higher R@K, lower MdR and MnR indicates better performance.

\subsection{Implementation Details}
The visual encoder, tag and text encoder of our TABLE model are initialized by the pretrained CLIP (ViT-B/32) \cite{radford2021learning} model. 
The parameters of tag and text encoder are shared in the transformer blocks but individually learned in the linear projection layer. 
The position embedding and the blocks of the cross-modal encoders are initialized by the first four layers of the CLIP’s text encoder. 
The initial learning rate is 1e-7 for visual and text encoder, and 1e-4 for cross-modal encoders. 
The TABLE model is trained for 5 epoches by Adam optimizer and warmup scheduler. The batch size is 128, except on DiDeMo is 48.
For MSR-VTT, MSVD and LSMDC, the max token length of caption and tag, and the frame length are set to 32, 32, 12, respectively. For DiDeMo, the value of the above parameters are 64, 32, 32, respectively.
To prevent the tag information from being overwhelmed by a larger number of video frames on DiDeMo,  we adopt all token output of the tag encoder instead of the [EOS] output as the input of \textbf{TG} cross-modal encoder. 
We adopt 5 pretrained experts (object, person, scene, motion and audio) to extract multi-modal tags on MSR-VTT and DiDeMo dataset, and 4 experts (without audio) on MSVD and LSMDC dataset.

\begin{table}[t]
	\centering
	\resizebox{!}{16mm}{
		\begin{tabular}{ccc|ccccc}
			\toprule
			\ & \  & \ & \multicolumn{5}{c}{Text-to-Video (T2V)} \\
			\midrule
			Type & Method & Pre-D & R@1 & R@5 & R@10 & MdR & MnR \\
			\midrule
			NO-CLIP & VSE \cite{kiros2014unifying} &- & 12.3 & 30.1 & 42.3 & 14.0 & -  \\
			NO-CLIP & CE &- & 19.8 & 49.0 & 63.8 & 6.0 & 23.1  \\
			NO-CLIP & SSML \cite{amrani2020noise} &HowTo100M & 20.3 & 49.0 & 63.3 & 6.0 & -  \\
			NO-CLIP & SUPPORT-SET &-  & 28.4 & 60.0 & 72.9 & 4.0 & -   \\
			NO-CLIP & FROZEN &WebVid-2M & 33.7 & 64.7 & 76.3 & 3.0 & - \\
			\hline\hline
			CLIP-based & CLIP &WIT& 37.0 & 64.1 & 73.8 & 3.0 & -  \\
			CLIP-based & CLIP4Clip-seqTransf  &WIT& 45.2 & 75.5 & 84.3 & 2.0 & 10.3  \\
			CLIP-based & CLIP4Clip-meanP &WIT & 46.2 & 76.1 & 84.6 & 2.0 & 10.0 \\
			CLIP-based & CLIP2VIDEO &WIT & 47.0 & 76.8 & 85.9 & 2.0 & 9.6  \\
			CLIP-based & CAMoE  &WIT & 46.9 & 76.1 & 85.5 & - & 9.8 \\
			CLIP-based & \textbf{TABLE} &WIT& 49.9 & 79.3 & 87.4 & 2.0 & \bf{9.1} \\
			\hline\hline
			CLIP-based & CAMoE$\ast$ &WIT & 49.8 & 79.2 & 87.0 & - & 9.4 \\
			CLIP-based & \textbf{TABLE$\ast$ (ours)} &WIT & \bf{52.3} & \bf{80.5} & \bf{87.9} & \bf{1.0} & 9.8 \\
			\bottomrule
	\end{tabular}}
	\captionsetup{font={small}}
	\caption{Retrieval results on the MSVD dataset. Pre-D represents Pretrained Dataset (as in subsequent tables).} 
	\label{tab3}
\end{table}

\begin{table}[t]
	\centering
	\resizebox{!}{19mm}{
		\begin{tabular}{ccc|ccccc}
			\toprule
			\ & \  & \ & \multicolumn{5}{c}{Text-to-Video (T2V)} \\
			\midrule
			Type & Method & Pre-D & R@1 & R@5 & R@10 & MdR & MnR \\
			\midrule
			NO-CLIP & JSFusion &- & 9.1 & 21.2 &34.1 & 36.0 & - \\
			NO-CLIP & CE &- & 11.2 & 26.9 & 34.8 & 25.3 & 96.8 \\
			NO-CLIP & HIT-Pretrained &HowTo100M & 14.0 & 31.2 & 41.6 & 18.5 & -  \\
			NO-CLIP & FROZEN &WebVid-2M & 15.0 & 30.8 & 39.8 & 20.0 & -  \\
			\hline\hline
			CLIP-based & CLIP &WIT& 11.3 & 22.7 & 29.2  & 56.5 & -  \\
			CLIP-based & MDMMT &WIT & 18.8 & 38.5 & 47.9 & 12.3 & 58.0 \\
			CLIP-based & CLIP4Clip-meanP &WIT & 20.7 &38.9 & 47.2 & 13.0 & 65.3 \\
			CLIP-based & CLIP4Clip-seqTransf &WIT & 22.6 & 41.0 & 49.1 & 11.0 & 61.0  \\
			CLIP-based & CAMoE &WIT & 22.5 & 42.6 & 50.9 & - & 56.5  \\
			CLIP-based & \textbf{TABLE (ours)} &WIT & 24.3 & 44.9 & 53.7 & 8.0 & 52.7 \\
			\hline\hline
			CLIP-based & CAMoE$\ast$ &WIT & 25.9 & \bf{46.1} & 53.7 & - & 54.4  \\
			CLIP-based & \textbf{TABLE$\ast$ (ours)} &WIT & \bf{26.2} & 45.9 & \bf{55.0} & \bf{7.0} & \bf{51.0} \\
			\bottomrule
	\end{tabular}}
	\captionsetup{font={small}}
	\caption{Retrieval results on the LSMDC dataset.}
	\label{tab4}
\end{table}

\begin{table}[t]
	\centering
	\resizebox{!}{15mm}{
		\begin{tabular}{ccc|ccccc}
			\toprule
			\ & \  & \ & \multicolumn{5}{c}{Text-to-Video (T2V)} \\
			\midrule
			Type & Method & Pre-D & R@1 & R@5 & R@10 & MdR & MnR \\
			\midrule
			NO-CLIP & S2VT \cite{venugopalan2014translating} &HowTo100M  & 11.9 & 33.6 & - & 13.0 & - \\
			NO-CLIP & FSE \cite{zhang2018cross} &Kinetics \cite{kay2017kinetics} & 13.9 & 36.0 & - & 11.0 & - \\
			NO-CLIP & CE &- & 16.1 & 41.1 & - & 8.3 & 43.7 \\
			NO-CLIP & TT-CE \cite{croitoru2021teachtext} &- & 21.6 & 48.6 & 62.9 & 6.0 & -  \\
			NO-CLIP & FROZEN &WebVid-2M & 34.6 & 65.0 & 74.7 & 3.0 & - \\
			\hline\hline
			CLIP-based & ClipBERT \cite{lei2021less} &COCO \cite{chen2015microsoft} + & 20.4 & 48.0 & 60.8 & 6.0 & - \\
			\ & \ & VGC \cite{krishna2017visual} & \ & \ &\ & \ & \  \\
			CLIP-based & CLIP4Clip-seqTransf &WIT & 42.8 & 68.5 & 79.2 & 2.0 & 18.9  \\
			CLIP-based & CLIP4Clip-meanP &WIT & 43.4 & 70.2 & 80.6 & 2.0 & 17.5  \\
			CLIP-based & \textbf{TABLE (ours)} &WIT & 47.9 & 74.0 & 82.1 & 2.0 & \bf{14.3} \\
			\hline\hline
			CLIP-based & CAMoE$\ast$ &WIT & 43.8 & 71.4 & 79.9 & 2.0 & 16.3  \\
			CLIP-based & \textbf{TABLE$\ast$ (ours)} &WIT & \bf{49.1} & \bf{75.6} & \bf{82.9} & \bf{2.0} & 14.8 \\
			\bottomrule
	\end{tabular}}
	\captionsetup{font={small}}
	\caption{Retrieval results on the DiDeMo dataset.}
	\label{tab5}
\end{table}

\begin{table}[t]
	\centering
	\resizebox{0.47\textwidth}{0.033\textheight}{
		\begin{tabular}{c|ccccc|ccccc|ccccc}
			\toprule
			\ &  \multicolumn{5}{c|}{Tag Modalities} & \multicolumn{5}{c|}{Text-to-Video (T2V)} &  \multicolumn{5}{c}{Video-to-Text (V2T)}\\
			\midrule
			Method & Object & Person & Motion & Scene & Audio & R@1 & R@5 & R@10 & MdR & MnR & 
			R@1 & R@5 & R@ 10  & MdR & MnR \\
			\midrule
			Baseline & - & - \ & - \ & - \ & - & 45.2 & 72.0 & 81.3 & 2.0 & 14.3 & 
			45.4 & 73.8 & 83.8 & 2.0 & 11.5 \\
			TABLE & \ding{52} & - & - & - & - & 45.6 & 72.8 & 81.7 & 2.0 & 13.8 & 45.3 & 73.6 & 83.9 & 2.0 & 11.1 \\
			TABLE & \ding{52} & \ding{52} & - & - & - & 45.6 & 72.8& 81.9 & 2.0 & 14.1 & 45.9 & 74.9 & 83.4 & 2.0 & 11.1 \\
			TABLE & \ding{52} & \ding{52} & \ding{52}  & - & - & 46.2 & 73.0 & 82.2 & 2.0 & \bf{13.8} & 45.9 & 74.7 & 83.8 & 2.0 & \bf{11.0} \\
			TABLE & \ding{52} & \ding{52} & \ding{52}  & \ding{52}  & -  & 46.5 & 73.4 & 82.2 & 2.0 & 13.9 & 
			45.6 & 74.8 & 83.6 & 2.0 & 11.2 \\
			TABLE &  \ding{52} & \ding{52} & \ding{52}  & \ding{52}  & \ding{52}  & \bf{46.8} & \bf{73.5} & \bf{82.2} & \bf{2.0} & 13.9 & 
			\bf{45.9} & \bf{75.1} & \bf{84.1} & \bf{2.0} & 11.1 \\
			\bottomrule
	\end{tabular}}
	\captionsetup{font={small}}
	\caption{Effects of multi-modal tags on MSR-VTT dataset.}
	\label{tab1}
\end{table}

\begin{table}[b]
	\centering
	\resizebox{0.47\textwidth}{0.035\textheight}{
		\begin{tabular}{c|cc|ccccc|ccccc}
			\toprule
			\ &  \multicolumn{2}{c|}{Task} & \multicolumn{5}{c|}{Text-to-Video (T2V)} &  \multicolumn{5}{c}{Video-to-Text (V2T)}\\
			\midrule
			Method & VTM & MLM  & R@1 & R@5 & R@10 & MdR & MnR & 
			R@1 & R@5 & R@ 10  & MdR & MnR \\
			\midrule
			Baseline & - & - \  & 45.2 & 72.0 & 81.3 & 2.0 & 14.3 & 
			45.4 & 73.8 & 83.8 & 2.0 & 11.5 \\
			TABLE &  - & - & 46.8 & 73.5 & 82.2 & 2.0 & 13.9 & 
			45.9 & 75.1 & 84.1 & 2.0 & 11.1 \\
			TABLE & \ding{52} & - & 46.7 & 73.9 & 82.1 & 2.0 & 13.6 & 
			46.6 & 75.0 & \bf{84.7} & 2.0 & \bf{10.6} \\
			TABLE & \ding{52} & \ding{52} & \bf{47.1} & \bf{74.3} & \bf{82.9} & \bf{2.0} & \bf{13.4} &\bf{47.2} & 74.2 & 84.2 & \bf{2.0} & 11.0 \\
			\bottomrule
	\end{tabular}}
	\captionsetup{font={small}}
	\caption{Effects of VTM and MLM on MSR-VTT dataset.}
	\label{tab6}
\end{table}

\subsection{Comparisons to the State of The Art}
In this subsection, we compare our model with state-of-the-art methods on four representative benchmarks. We divide existing methods into CLIP-based and NO-CLIP, where CLIP-based methods usually have better performance.

As shown in Table \ref{tab2}, without inference strategy, our TABLE model achieves great performance on the MSR-VTT dataset. For example, our method surpasses CAMoE \cite{cheng2021improving} by a large margin of 2.5 and 2.1 on R@1 in text-to-video (T2V) and video-to-text (V2T) task respectively.
It not only demonstrates the importance of multi-modal information, but also proves the superiority of our tagging method for enabling cross-modal alignment. Furthermore, with the inference strategy, our method achieves SOTA performance, i.e., 52.3 R@1 in T2V and 51.8 R@1 in V2T task.
For the MSVD dataset,  as illustrated in Table \ref{tab3}, our TABLE method surpasses CAMoE by 3.0 on R@1 in T2V task without inference strategy, and finally achieves SOTA performance of 52.3 R@1.
We believe that on scale-limited datasets, the explicit guidance of multi-modal tags is more important for visual-text alignment.
LSMDC dataset contains the most videos and each video matches only one caption, and thus most methods perform pooly. But our method still improves 1.8 on R@1 and achieves SOTA performance (26.2 R@1) in T2V task, as shown in Table \ref{tab4}.
Finally, for the video-paragraph retrieval task on DiDeMo, our method obtains 5.3 R@1 gains over SOTA method CAMoE, as presented in Table \ref{tab5}. The alignment of long videos and long text meets greater challenge compared to short video-text pairs, and thus the introduction of multi-modal tags shows greater advantages on this dataset. 
In conclusion, our TABLE model proves to be effective and superior by achieving SOTA performance on various benchmarks.

\subsection{Abalation Studies}
\subsubsection{Effects of Multi-Modal Tags.}
The multi-modal information of the video are fully exploited in this work by an explicit way of tagging.
In this subsection, we delve into the impact of multi-modal tags on model's performance. 

As shown in Table \ref{tab1}, our backbone model achieves R@1 of 45.2 for text-to-video (T2V) retrieval, which is already a comparable performance on the MSR-VTT dataset. The R@1 of T2V is improved to 45.6 after the object tags are introduced by our TABLE model, and the R@1 of V2T is further improved to 45.9 with object and person tags. 
It proves that the visual-text alignment benefits from additional tag information.
The object and person tags contain local information, which can guide the local attention of visual features. 
But in video retrieval task, the motion modality plays an important role.
With motion tags, the performance is obviously improved, i.e., the T2V R@1 reaches 46.2. It is further promoted to 46.5 after introducing scene tags, which usually describe the background information.
Finally, the T2V R@1 is improved to 46.8 with audio tags, which contain information not included in visual features, which is helpful for the retrieval of videos like news report.
To sum up, introducing multi-modal tags is beneficial for video-text retrieval task.

\begin{figure*}[ht]
	\centering
	\includegraphics[width=0.8\linewidth]{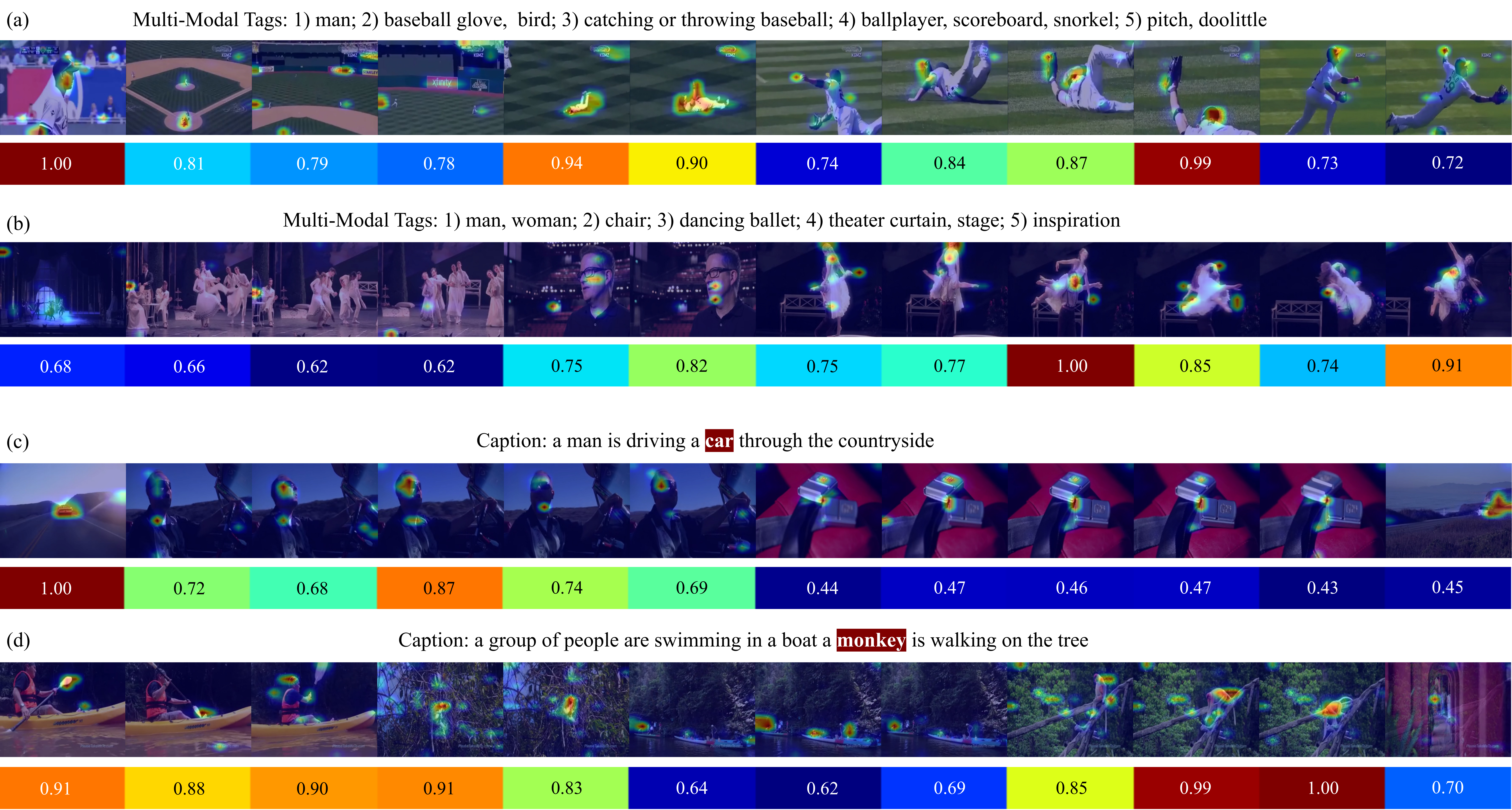}
	\captionsetup{font={small}}
	\caption{Visualization results of TABLE. We visualize the spatial attention in each frame, where brighter regions represent higher spatial attention. And, we visualize the temporal attention through the color bar below the frame list, where redder color represents higher attention. }
	\label{fig3}
\end{figure*}

\subsubsection{Effects of VTM and MLM.}
VTM and MLM are proposed as auxiliary supervisions with a joint cross-modal encoder, both for facilitating the video-text interaction.

As shown in Table \ref{tab6}, with the help of VTM, the video-to-text retrieval performance is improved from 45.9 to 46.6 at R@1. Using both VTM and MLM, the model performance is further promoted. 
The R@1 of text-to-video reaches 47.1, and the R@1 of video-to-text achieves 47.2. Although VTM is a binary classification task, the model can only make well judgment on hard negative with profound understanding of video and text content. MLM is a more complex task which requires fine-grained alignment between the visual features and text tokens for inferring masked words. 
In general, these two tasks strengthen the interaction between video and text, and thus effectively improves the model performance.

\subsection{Visual Analysis}
\label{s4.4}
To clearly reveal the anchor role of multi-modal tags, we present some visualization results of the \textbf{TG} cross-modal encoder and the joint cross-modal encoder using the Attention Rollout \cite{abnar2020quantifying} method.

In Fig. \ref{fig3}, we simultaneously visualize the spatial attention and temporal attention, where the temporal attention is calculated by the cross attention between frame features and the overall representation of multi-modal tags (or specific word token of the caption). Higher temporal attention indicates higher weights in representing the whole video. 
In Fig. \ref{fig3}(a) and Fig. \ref{fig3}(b), the multi-modal tags are treated as an entirety for calculating cross attention with visual features.
As shown, the cross attention focuses on some frames that highly correlated with multi-modal tags.
For example, in Fig. \ref{fig3}(a), the 1st frame (highly correlated with ``man'' and ``ballplayer'') and the 10th frame (highly correlated with ``catching or throwing baseball'' and ``baseball glove'') are highlighted, while the 2rd-4th frames describing distant view of the stadium are suppressed. 
For each frame, important spatial regions like head, body, baseball glove are getting more attention than unimportant backgrounds.
Although the multi-modal tags might contain noise, e.g., ``bird'' in Fig. \ref{fig3}(a), they can still guide the model to pay more attention to key frames and key regions. We also find that the motion modality show dominant role in the cross-modal attention.
For example, in Fig. \ref{fig3}(b), the 9th and 12th frames describing standard ``dancing ballet'' are highlighted, while the first few frames depicting the scene and person are suppressed. 
To sum up, the \textbf{TG} cross-modal encoder is capable of selecting key frames and key regions from redundant visual features with the guidance of multi-modal tags, which is beneficial for accurate video-text retrieval. 

To further investigate the capabilities of our cross-modal encoder, we also visualize the cross attention in the joint cross-modal encoder. In detail, here we visualize the cross attention between visual features and specific word token of the caption rather than the overall representation. 
For example, in Fig. \ref{fig3}(c), given the ``car'' word, the 1st frame is highlighted as it gives the whole picture of the car. The spatial attention also focuses on the car body.
The last few frames are judged to be less correlated as they just focus on seat belts. 
In Fig. \ref{fig3}(d), the 10th and 11th frames are determined as the most relevant frames to the word ``monkey'' because they describe the monkey with a close view. The other frames describing about ``people'' or ``swiming'' are suppressed.
Thus, it can be concluded that the proposed joint cross-modal encoder is capable of modeling the fine-grained correlation between video and a single word of text, which is inherited from VTM and MLM tasks. Moreover, it reveals the strong capability of \textbf{TG} cross-modal encoder as the parameters are shared between it and the joint cross-modal encoder.

\section{Conclusion}
In this paper, we propose to integrate multi-modal tags as anchors to motivate the video-text alignment. Specifically, we construct a TABLE model to jointly encode multi-frame visual features and multi-modal information (including object, person, scene, motion and audio).
To further enhance the video-text interaction, we introduce VTM and MLM tasks on the triplet input of [visual, tag, text] as auxiliary supervisions.
Our proposed TABLE model achieves SOTA performance on various benchmarks, including MSR-VTT, MSVD, LSMDC and DiDeMo, which indicates the superiority of introducing multi-modal tags as anchors for video-text retrieval task. The ablation studies and visualization results further reveal the anchor role of multi-modal tags in guiding visual-text alignment.

\vspace{.2em}
\bibliography{aaai23}

\end{document}